\pdfoutput=1

\documentclass[11pt]{article}

\usepackage[preprint]{acl}

\usepackage{times}
\usepackage{latexsym}

\usepackage[T1]{fontenc}

\usepackage[utf8]{inputenc}

\usepackage{microtype}

\usepackage{inconsolata}

\usepackage{graphicx}

\usepackage{algorithm}
\usepackage{algpseudocode}
\usepackage{amsmath,amsfonts,amssymb}
\usepackage{booktabs}

\usepackage{graphicx}
\usepackage{paralist}
\usepackage{xcolor}
\usepackage{xspace}
\usepackage{adjustbox}
\usepackage{tabularx}
\usepackage{multirow}
\usepackage[flushleft]{threeparttable}
\usepackage{diagbox} 
\usepackage{colortbl}
\usepackage{xcolor}
\usepackage{pict2e}
\usepackage{hyperref} 

\title{Enhancing Frame Detection with Retrieval Augmented Generation}

\newsavebox{\ORCIDlogo}
\savebox{\ORCIDlogo}{%
\setlength{\unitlength}{\dimexpr 1em/256\relax}%
\begin{picture}(256,256)%
  \color[HTML]{A6CE39}\put(128,128){\circle*{256}}%
  \color{white}%
  \put(78.6,199.2){\circle*{20}}%
  \moveto(70.9,176,9)\lineto(86.3,176,9)\lineto(86.3,69.8)\lineto(70.9,69.8)%
  \closepath\fillpath%
  \moveto(108.9,176.9)\lineto(150.5,176.9)%
  \curveto(190.1,176.9)(207.5,148.6)(207.5 ,123.3)%
  \curveto(207.5,95,8)(186,69.7)(150.7,69.7)%
  \lineto(108.9,69.7)%
  \closepath\fillpath%
  \color[HTML]{A6CE39}%
  \moveto(124.3,83.6)\lineto(148.8,83.6)%
  \curveto(183.7,83.6)(191.7,110.1)(191.7,123.3)%
  \curveto(191.7,144.8)(178,163)(148,163)%
  \lineto(124.3,163)%
  \closepath\fillpath%
\end{picture}%
}
\newcommand\orcidicon[1]{\href{https://orcid.org/#1}{\usebox{\ORCIDlogo}}}

\author{%
    Papa Abdou Karim Karou Diallo\orcidicon{0000-0002-0614-5185} \\  
    Mila, Polytechnique Montréal \\  
    \texttt{diallokarou28@polymtl.ca}  
    \And  
    Amal Zouaq\orcidicon{0000-0002-4791-0752} \\  
    Mila, Polytechnique Montréal \\  
    \texttt{amal.zouaq@polymtl.ca}  
}

\begin{document}
\maketitle
\begin{abstract}
Recent advancements in Natural Language Processing have significantly improved the extraction of structured semantic representations from unstructured text, especially through Frame Semantic Role Labeling (FSRL). Despite this progress, the potential of Retrieval-Augmented Generation (RAG) models for frame detection remains under-explored. In this paper, we present the first RAG-based approach for frame detection called \textbf{RCIF} (\textbf{R}etrieve \textbf{C}andidates and \textbf{I}dentify \textbf{F}rames). RCIF is also the first approach to operate without the need for explicit target span and comprises three main stages: (1) generation of frame embeddings from various representations ; (2) retrieval of candidate frames given an input text; and (3) identification of the most suitable frames. 
We conducted extensive experiments across multiple configurations, including zero-shot, few-shot, and fine-tuning settings. 
Our results show that our retrieval component significantly reduces the complexity of the task by narrowing the search space thus allowing the frame identifier to refine and complete the set of candidates. Our approach achieves state-of-the-art performance on FrameNet 1.5 and 1.7, demonstrating its robustness in scenarios where only raw text is provided. Furthermore, we leverage the structured representation obtained through this method as a proxy to enhance generalization across lexical variations in the task of translating natural language questions into SPARQL queries.
\end{abstract}

\begin{figure}[t!]
\centering
\includegraphics[width=0.5\textwidth]{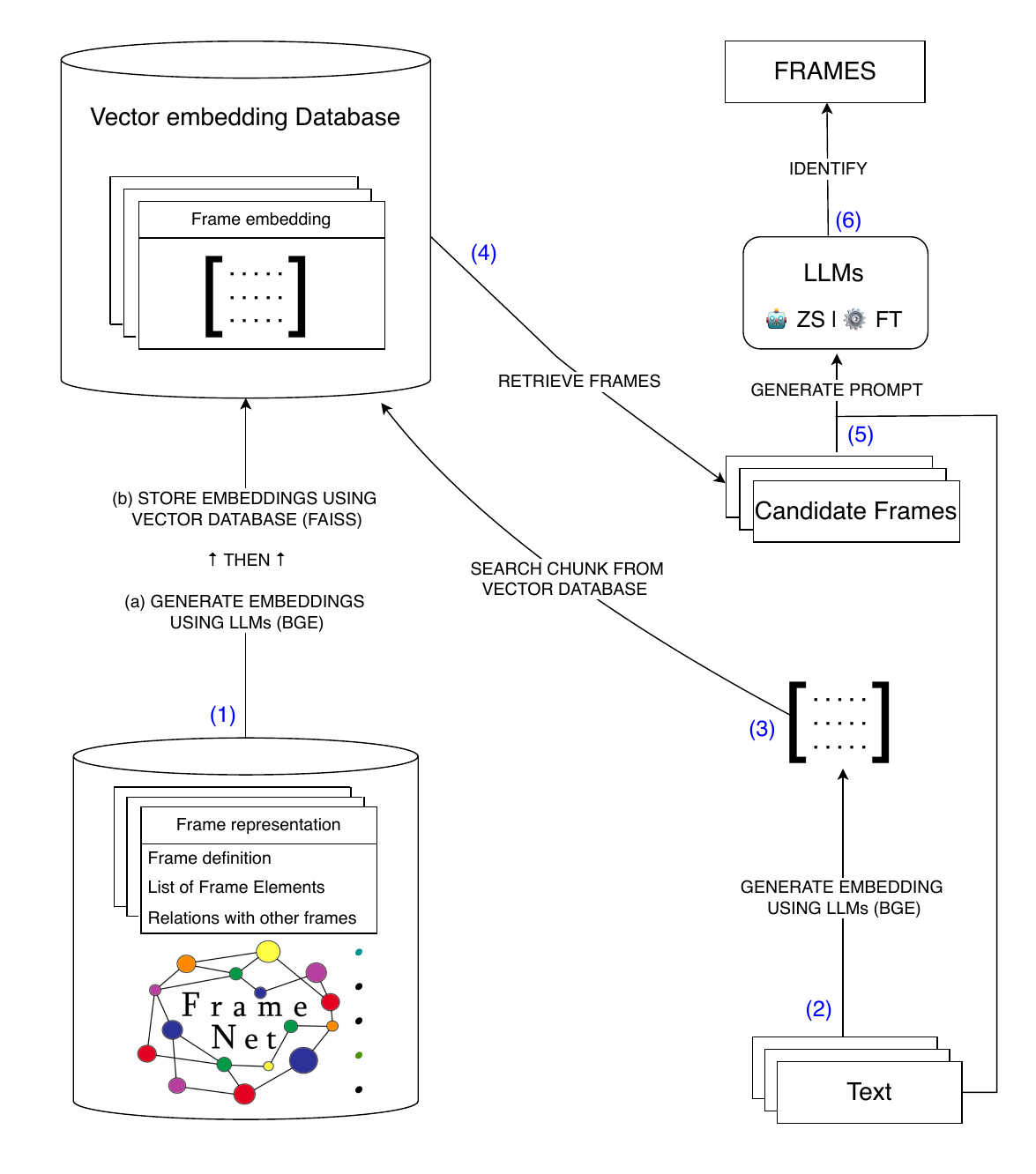}
\caption{Overview of our proposed method called \textbf{RCIF} (\textbf{R}etrieve \textbf{C}andidates and \textbf{I}dentify \textbf{F}rames). (1) Frame embeddings are generated using an embedding model based on various frame representations. These embeddings are stored in a vector database. (2-3) Given an input text, the system retrieves candidate frames based on similarity scores of input text and frames embeddings. (4-5) An LLM is then fine-tuned with dynamic prompts to select the best matching frames from the retrieved candidates, completing identification process.}
\label{fig:pipeline}
\end{figure}

\section{Introduction}
Large language models (LLMs) have significantly contributed to better semantic representations of natural language sequences, as witnessed by their state-of-the art performance in several natural language understanding tasks. However, LLMs are prone to a degradation of performance depending on their natural language input or prompt, which leads to limitations in terms of robustness and generalization \citep{leidinger2023language, mengsfr, zheng2023prompt}. Our high-level objective in this work is to explore if structured representations of natural language input, and more precisely frame semantics, can contribute to more robust models. Another objective is to explore a RAG-based approach for frame detection.
Frames, as defined in Frame Semantics \citep{baker1998berkeley}, are conceptual structures that capture the semantic and syntactic relationships underlying language. They are helpful in providing a structured semantic context for understanding relationships between entities, enabling tasks like Machine Reading Comprehension \citep{guo2020incorporating, flanigan2022meaning, bonn2024meaning} and Information Extraction \citep{su2021knowledge, li2024comprehensive, chanin2023open, su2023span} to be more accurate and contextually aware.  Each frame represents a specific type of situation, event, or object along with the roles (or frame elements) associated with it.
Frame identification involves selecting the most suitable semantic frame for a target word in a sentence context. 
Generally, state-of-the-art approaches require to indicate, in a free-form text, the explicit target for which a frame should be found. Consequently, they struggle to generalize to text without explicit targets, limiting their utility in broader downstream tasks, where these targets are not necessarily known.

To solve these aforementioned limitations, our proposed method called \textbf{RCIF} (\textbf{R}etrieve \textbf{C}andidates and \textbf{I}dentify \textbf{F}rames) for frame detection, as depicted in Figure \ref{fig:pipeline}, leverages a RAG framework combined with a generative large language model (LLM) functioning mostly as a classifier. The approach operates in three main stages. First, we generate embeddings for candidate frames based on a variety of representations, including frame labels, descriptions, lexical units (LUs), and frame elements (FEs), and store these embeddings in a vector database. Next, for a given text input, we retrieve frames with the highest similarity scores as candidate frames. Finally, a generative LLM selects the best-matching frames from these candidates, effectively acting as a classifier. Given the hypothesis that structured representations have the potential to bridge the gap between different lexical variations of the same text, we utilize frames as an intermediate structured representation to translate natural language questions into SPARQL queries. 

The contributions of this paper are as follows: 
\begin{enumerate}
    \item We present the first approach that leverages a RAG model for frame detection. 
    \item We also introduce the first method to perform target-free frame detection directly from text. This makes our framework more useful for any text that does not have a predefined target.
    \item Finally, we demonstrate that the structured representations obtained through this approach has the potential of improving the generalization of SPARQL query generation for different lexical variations of a question. 
\end{enumerate}

Overall, our approach not only enhances frame detection accuracy but also addresses the challenge of target-free input by dynamically narrowing the search space to relevant frames, improving both recall and precision in frame identification.

\section{Related Works}
\subsection{Frame semantic parsing}
The literature on frame parsing can be broadly divided into two main approaches: methods that frame the task as a sequence-to-sequence (Seq2Seq) generation problem and methods based on representation learning. 

Seq2Seq approaches \citep{sutskever2014sequence,raffel2020exploring,kalyanpur2020open,chanin2023open} define frame parsing as a generative task, decomposing it into subtasks such as trigger identification, frame classification, and argument extraction. These methods leverage pre-trained language models and task-specific optimizations to balance the subtask distributions and mitigate data scarcity \citep{kalyanpur2020open,chanin2023open}. As illustration, models like T5 \citep{raffel2020exploring} are pre-trained on PropBank \citep{kingsbury2002treebank, kingsbury2003propbank} and FrameNet exemplars, with text augmentation techniques applied to improve robustness and FrameNet lexical units incorporated to enhance frame classification accuracy. Then, they are fine-tuned on each of the aforementioned sub-tasks. 
For exemple, \citep{kalyanpur2020open} adopt a shared encoder with specialized decoders for each sub-task, enabling the model to leverage a common representation while handling each task independently within the same architecture. The Seq2Seq methods share a focus on utilizing the flexibility of generative models to capture the sequential nature of frame parsing tasks. 

Representation learning approaches, in contrast, focus on constructing enriched embeddings that align sentence-level context (or just the target span) with candidate frames \citep{hartmann2017out, jiang2021exploiting}. They often employ graph-based techniques, such as Graph Neural Networks (GNNs) \citep{wu2020comprehensive}, or contrastive learning \citep{ju2024towards} to incorporate external knowledge and enhance the robustness of frame representations. These methods also emphasize semantic alignment through embedding techniques that integrate knowledge from FrameNet's structure. Graph-based methods, for instance, exploit relationships between frames and frame elements \citep{su2021knowledge, zheng2022double, tamburini2022combining}, while contrastive learning approaches align contextual representations of target span with frame embeddings to refine predictions  \citep{hartmann2017out,jiang2021exploiting,an2023coarse}.

A key limitation that hinders these approaches from generalizing effectively and being applicable in real-world scenarios is their reliance on both the context (text/sentence) and a target, which has to be specified at input. For instance, consider the sentence: \textit{"We help people train for and find jobs that make it possible for them to get off of welfare."}. To detect the frame \textit{"Assistance,"} these approaches require information about the position of the target span 
\textit{"help"}. This dependency restricts their flexibility and reduces their practical utility in settings where only raw text without predefined targets is available.

To address the computational challenges in frame detection, we propose an approach to reduce the search space by first retrieving a subset of potential candidate frames that are likely to be evoked by the sentence. By limiting the frame search space, we aim to decrease the number of frame evaluations for each word, thereby reducing overall complexity.

\subsection{SPARQL queries generation}
One interesting task that can benefit from the same semantic representation for diverse natural language formulations is the task of SPARQL query generation from natural language questions. The existing literature on SPARQL query generation from natural language predominantly centers around the use of pretrained language models \citep{diallo2024comprehensive, reyd2023assessing, banerjee2022modern, emonet2024llm, zahera2024generating}. Small Language Models (SLMs) are commonly fine-tuned in an end-to-end manner, often incorporating mechanisms such as copy strategies to minimize errors in generating URIs within the final queries \citep{diallo2024comprehensive, banerjee2022modern}. In contrast, Large Language Models (LLMs) are partially employed to generate answers directly by grounding facts from knowledge bases, bypassing the intermediate step of explicit query generation \citep{shavarani2024entity, alawwad2024enhancing, muennighoff2022sgpt}. Another line of research explores various prompt engineering techniques during fine-tuning, with or without demonstrations, and sometimes incorporates additional contextual information such as the URIs required in the generated SPARQL queries \citep{diallo2024comprehensive, muennighoff2022sgpt, luo2023chatkbqa}. Notable examples include models such as Code Llama v2 7B \citep{roziere2023code}, Mistral 7B v0.3 \citep{jiang2023mistral7b}, and Mistral 7B Instruct v0.3\footnote{https://huggingface.co/mistralai/Mistral-7B-Instruct-v0.3}. Despite leveraging large pretrained models, these approaches often neglect a detailed examination of their performance on various formulations of the same question, or try to include paraphrases to produce more robust models. A simple comparison between performance on template questions and reformulated questions in the LCQuAD 2 dataset shows that natural questions are poorly handled by models \citep{diallo2024comprehensive, reyd2023assessing}.   

By leveraging frame semantic parsing to represent natural language questions as structured representations, our approach seeks to enhance generalization on template-free or reformulated questions.

\section{Methodology}
\subsection{Datasets}
To ensure a fair comparison with prior work, we assess our model's performance on FrameNet 1.5, adhering to the original train/dev/test data splits established by \citep{das2014frame}. Additionally, we extend our evaluation to FrameNet 1.7, released in 2016, which offers approximately 20\% more gold-standard annotations compared to FrameNet 1.5. For both FrameNet 1.5 and FrameNet 1.7, we follow the data splits defined by An et al. \citep{an2023coarse}. Table \ref{tab:dataset_distribution_1} provides details on the number of examples in each split, including the exemplars dataset commonly used for pretraining.

\begin{table}
    \centering
    \caption{Dataset Distribution for FrameNet 1.5 and FrameNet 1.7}
    \begin{adjustbox}{width=\linewidth}
    \begin{tabular}{lcccccc}
        \toprule
        \multirow{2}{*}{\textbf{Dataset Split}} & \multicolumn{2}{c}{\textbf{FrameNet 1.5}} & & \multicolumn{2}{c}{\textbf{FrameNet 1.7}} \\
        \cmidrule(lr){2-3} \cmidrule(lr){5-6}
         & \textbf{All} & \textbf{Uniques} & & \textbf{All} & \textbf{Uniques} \\
        \midrule
        Train & 16,621 & 2,653 & & 19,391 & 3,353 \\
        Dev & 2,284 & 326 & & 2,272 & 326 \\
        Test & 4,428 & 875 & & 6,714 & 1,247 \\
        Exemplars & 153,946 & 147,483 & & 192,431 & 168,266 \\
        \midrule
        \textbf{\# Frames} & \multicolumn{2}{c}{1,019} & & \multicolumn{2}{c}{1,221} \\
        \bottomrule
    \end{tabular}
    \end{adjustbox}
    \label{tab:dataset_distribution_1}
\end{table}

\subsection{Concepts definition and task description}
\label{sec:methodology-task-description}
A FrameNet frame \(f\) is defined by its label, representing its name, and accompanied by a textual description that provides a comprehensive explanation of its semantics. Each frame is further characterized by a set of frame elements (FEs), which serve as its core semantic components, capturing contextual and relational information about the frame. Additionally, frames are associated with a set of lexical units (LUs), consisting of lemmas paired with their parts of speech, which denote the frame or specific aspects of it. Within a sentence, tokens (words or phrases) that evoke a frame are referred to as targets.

We formulate our task as the detection of frames \(f_i, \dots, f_j\) within a sentence \(S = w_1, \dots, w_n\), without any predefined target span for frame triggering. This formulation reflects real-world applications, where the input typically consists of raw text, and frames must be identified directly from it. Unlike approaches such as CoffTea \citep{an2023coarse} and related works that rely on both a sentence and a specific target span \(t = w_{t_s}, \dots, w_{t_e}\) (with \(w_{t_s}\) and \( w_{t_e}\) respectively corresponding to the start and the end of the target span) to identify a single frame, our model, \textbf{RCIF}, is designed to identify frames in the absence of such target span information. 

The original dataset consists of entries such as the sentence \textit{ "I was sad when I could n't go to the snack bar to \underline{buy} a soda."} where the underlined span is the target whose position is provided 
as well as the frame  which is "Commerce\_buy" and its definition. Thus, the entry is repeated as many times as there are target spans or possible frames for the same sentence with, for every instance, a new target span and a new frame. 

To generalize over target-less raw text, we adapt the original dataset by grouping frames by sentence so that each unique sentence is associated with all the frames it evokes. For instance, the six occurrences of the previous sentence are merged into a single entry, discarding information about the target and retaining only the following list of possible frames such as "Emotion\_directed", "Capability", "Likelihood", "Locative\_relation", "Goal", "Commerce\_buy" and "Temporal\_collocation". As one can see, this new dataset formulation constitutes a harder task than the previous one. We thus address this problem by first retrieving a set of candidates and then identifying the appropriate frames, as detailed in Section \ref{sec:methodology-retrieval-component}.

As shown in Table \ref{tab:dataset_distribution_2}, the resulting data distribution includes a minimum of 1 frame per sentence, a maximum of 24 frames per sentence, and an average of 5 frames per sentence for FrameNet 1.5. For FrameNet 1.7, we observe similar statistics, with a minimum of 1 frame, a maximum of 23 frames, and an average of 5 frames per sentence. The exemplar data presents an average and minimum of one frame per sentence across both datasets, with a maximum of 14 frames for FrameNet 1.5 and 21 frames for FrameNet 1.7.

\begin{table}
    \centering
    \caption{Number of frames per sentence in the different splits for FrameNet 1.5 and FrameNet 1.7.}
    \begin{adjustbox}{width=\linewidth}
    \begin{tabular}{lcccccc}
        \toprule
        \multirow{2}{*}{\textbf{Dataset Split}} & \multicolumn{3}{c}{\textbf{FrameNet 1.5}} & \multicolumn{3}{c}{\textbf{FrameNet 1.7}} \\
        \cmidrule(lr){2-4} \cmidrule(lr){5-7}
         & \textbf{Min} & \textbf{Max} & \textbf{Mean} & \textbf{Min} & \textbf{Max} & \textbf{Mean} \\
        \midrule
        Train & 1 & 21 & 5 & 1 & 21 & 5 \\
        Validation & 1 & 18 & 6 & 1 & 18 & 6 \\
        Test & 1 & 24 & 4 & 1 & 23 & 4 \\
        Exemplars & 1 & 14 & 1 & 1 & 21 & 1 \\
        \bottomrule
    \end{tabular}
    \end{adjustbox}
    \label{tab:dataset_distribution_2}
\end{table}

Our general architecture consists of three steps: (1) generation of frame embeddings from various representations (section \ref{sec:methodology-retrieval-component}); (2) retrieval of candidate frames given an input text (section \ref{sec:methodology-retrieval-component}); and (3) identification of the best frames (section \ref{sec:methodology-detection-component}). In our experiments, we employ the Llama 3.2-3B model \citep{dubey2024llama} model under the technical configuration depicted in Table \ref{tab:fine-tuning-technical-details}.

\subsection{Retrieval of candidates}
\label{sec:methodology-retrieval-component}

In this phase, we employ a frozen-RAG model to facilitate candidate retrieval. For the retrieval component, we compare different frame representations using labels, descriptions, lexical units (LUs), or frame elements (FEs).  
We systematically explore all these options, generating embeddings for each frame representation, which are then stored in a vector database, as illustrated in Figure \ref{fig:example_frames_representation} in Appendix \ref{sec:appendix-retrieval-component}. When processing a new text, we generate its vector embedding and perform a similarity search to retrieve the top \(k\) candidates based on similarity scores. For embedding generation, we utilize the English version of BGE\footnote{https://huggingface.co/BAAI} \citep{li2024makingtextembeddersfewshot,bge_embedding}. 
BGE is a unified embedding model designed to produce general-purpose text embeddings applicable to diverse tasks, including retrieval, ranking, and classification, across settings such as question answering and conversational systems. The model employs a BERT-like architecture, utilizing the hidden state of the [CLS] token as the embedding, and incorporates a comprehensive training recipe involving pre-training, general-purpose contrastive learning, and task-specific fine-tuning. 
Finally, we used FAISS to efficiently conduct the similarity search.


\subsection{Identification of frames}
\label{sec:methodology-detection-component}
Identifying the set of frames evoked by a sentence without a specified target span presents significant complexity, influenced by sentence length and the total number of frames in the lexicon (1019 in FrameNet 1.5 and 1221 in FrameNet 1.7) (see Figure \ref{fig:complexity_analysis_frame_detection} in Appendix \ref{sec:appendix-frame-detection-component}). To address this, we reduce the search space by generating a list of potential frame candidates that guide the model towards the correct frames. This is achieved through an initial retrieval component (described in section \ref{sec:methodology-retrieval-component}) that provides a list of candidates prior to using a pretrained LLM to finalize frame selection. The LLM is fine-tuned in an \textit{Instruction-Input-Output} format. This prompt design allows the fine-tuned model to not only gain inspiration from the retrieved candidates but also to consider frames identified in previous batches, thereby accommodating retrieval imperfections where some correct frames may not appear in the candidates list. More detail about the prompt and this process is provided in Appendix \ref{sec:appendix-frame-detection-component}.\\

\subsection{Frames for SPARQL query generation}
In this section, we investigate the role of structured representations in enhancing the generalization of SPARQL query generation across lexical variations of a question. As reported in \citep{diallo2024comprehensive}, template-based questions yield higher performance than reformulated natural language questions, indicating that transforming unstructured questions into structured representations has the potential to improve the accuracy and relevance of SPARQL query generation.
Thus, we fine-tuned several LLMs including Llama 3.1-8B\footnote{\url{https://huggingface.co/meta-llama/Llama-3.1-8B}}, Llama 3.2-3B model \citep{dubey2024llama}, Phi-4\footnote{\url{https://huggingface.co/microsoft/phi-4}} and Qwen2.5-7B-Instruct\footnote{\url{https://huggingface.co/Qwen/Qwen2.5-7B-Instruct}}. The core of the paper focuses on the best-performing results obtained with Llama 3.1-8B while the results of the other LLMs are provided in Appendix \ref{sec:appendix-additional-experiments-phi4-qwen}.

\paragraph{Datasets LCQ2F and LCQ2F+.}
\label{sec:datasets-lcq2f-and-lcq2f+-construction}
In order to get datasets of natural language questions that also incorporate frame semantic parsing, we perform the intersection of the datasets \textit{LC-QuAD 2.0} and \textit{WikiBank} that are described in Appendix \ref{sec:introduction-of-lcq2-and-wikibank}.
The resulting dataset \textit{LCQ2F} of size \(2,963\) entries (train: \(2,146\) | val: \(224\) | test: \(593\)) consists of the subset of \textit{LC-QuAD 2.0} \citep{LCQUAD2} questions remaining after their intersection with WikiBank \citep{sas2020wikibank} where each question is augmented by the list of frames evoked in it. The construction of \textit{LCQ2F} is based on the following postulate: given an \textit{LC-QuAD 2.0} question \textit{NLQ} and its corresponding SPARQL query \textit{Q}, along with a sentence \textit{S}  from WikiBank, \textit{NLQ} evokes the same frame as \textit{S}  if and only if \textit{Q} and \textit{S} share the same set of relations and classes from the Wikidata Knowledge Base. When this condition is met, the set of frames evoked in \textit{S} is attributed to \textit{NLQ} and subsequently used as augmented information.

The dataset \textit{LCQ2F+} extends \textit{LCQ2F} (with the same train, validation and test splits size) by incorporating additional information, including the textual descriptions of the frames and a detailed list of associated Frame-Elements. This provides richer structured representations to further contextualize the \textit{NLQ}. For each dataset the splits size double for \textit{Combined questions} (more details provided in Appendix \ref{sec:appendix-constructed-datasets}).

\section{Results}
\subsection{Retrieval component}

Table \ref{tab:retrieval-results} presents the performance of candidate frames retrieval using the English version of the BGE embedding model \citep{li2024makingtextembeddersfewshot,bge_embedding}. Retrieval is configured to select \(K = \text{max}_{\text{frames}} = 24\) candidates, which corresponds to the maximum number of frames a sentence might evoke. This setting is chosen to maximize recall, ensuring that the subsequent detection/identification stage has a high likelihood of finding relevant frames among the candidate set. 

As seen in the table, the third frame representation (\textit{Representation 3}, including frame label and description, the list of frame-elements and lexical-units, yields on average the highest recall, highlighting it as the most effective representation format. Across all datasets and frame representations, precision remains low, a consequence of maximizing recall by retrieving a surplus of frames (\( \text{max}_{\text{frames}} = 24 \)) compared to the average need (\( \text{avg}_{\text{frames}} = 5 \)). This retrieval strategy not only provides the frame detector component with the broadest possible set of relevant candidates, but it also encourages the model during fine-tuning to rely less on parametric memory and more on generalization, enhancing its robustness.

\begin{table}
    \caption{Retrieval Performances on FrameNet 1.5 and FrameNet 1.7 train sets (\%)}
    \centering
    \begin{threeparttable}
    \begin{adjustbox}{width=\linewidth}
    \begin{tabular}{lcccccc}
        \toprule
        & \multicolumn{2}{c}{\textbf{FrameNet 1.5 }} & \multicolumn{2}{c}{\textbf{FrameNet 1.7 }} & \multicolumn{2}{c}{\textbf{Average}} \\
        \cmidrule(lr){2-3} \cmidrule(lr){4-5} \cmidrule(lr){6-7}
        \textbf{Metrics} & \textbf{P} & \textbf{R} & \textbf{P} & \textbf{R} & \textbf{P} & \textbf{R} \\
        \midrule
        Rep1 & 4 & 79 & 3 & 72 & 4 & 76 \\
        Rep2 & 5 & 90 & 4 & 77 & 5 & 84 \\
        Rep3 & 5 & 89 & 4 & 81 & 5 & \textbf{85} \\
        \bottomrule
    \end{tabular}
    \end{adjustbox}
    \begin{tablenotes}
     \item[]R : Recall 
     \item[]P : Precision 
     \item[] Rep\(_i\)  : Representation\(_i\) with \(i \in \{1, 2, 3 \} \)
  \end{tablenotes}
    \end{threeparttable}
    \label{tab:retrieval-results}
\end{table}

\subsection{Frame detection}
\paragraph{Experiments. } We conducted multiple experiments using several models such as Llama 3.2-3B model \citep{dubey2024llama}, Llama 3.1-8B, Phi-4, and Qwen2.5-7B-Instruct across three main settings: zero-shot, few-shot, and fine-tuning but we just keep the best performing model (Llama 3.2-3B). For each setting, we implemented two configurations: one that explicitly indicates the number of frames the model is expected to detect, and another that does not. This distinction allows us to test the hypothesis that, without specifying the number of frames, the model may struggle to accurately determine the appropriate number of candidates to select.

Additionally, we included a baseline experiment that involves fine-tuning the model to generate frames without leveraging the retrieval component as an initial step.

\paragraph{Results of frame detection.}
Table \ref{tab:frame-detection-results-fn15-fn17} presents the frame detection results of the Llama 3.2 - 3B \citep{dubey2024llama} model fine-tuned for 10 epochs on the complete training sets of FrameNet 1.5 and FrameNet 1.7. While the best accuracy is reported in \citep{tamburini2022combining} and COFFTEA \citep{an2023coarse} for FrameNet 1.5, our model achieves higher precision/recall, outperforming prior work by approximately 4 points in recall. This  improvement is attributed to the reduction in search space during the retrieval phase and effective de-noising (elimination of irrelevant candidates) by the fine-tuned LLM. Starting with a retrieval phase precision of 5\% and a recall of 89\% as shown in Table \ref{tab:retrieval-results}, our final model enhances both metrics to around 92\%, indicating successful removal of incorrect candidates while retaining relevant ones. 
This effect is even more pronounced in FrameNet 1.7, where our model achieves top performance across all metrics, with a precision of 99\% and a recall of 97\%, demonstrating that the model effectively filters out incorrect candidates. The additional training samples in FrameNet 1.7 (26\% more than FrameNet 1.5) further contribute to this improvement. 
Interestingly, specifying the exact number of frames to be generated had minimal impact on performance, sometimes even reducing it. This finding is crucial for real-world applications, as providing exact frame counts is often infeasible with out-of-distribution data. Consistent with the task description in section \ref{sec:methodology-task-description}, this framework is designed for practical deployment where only the sentence is provided without target or frame count specifications.

\begin{table*}[!ht]
    \centering
    \caption{Performance (\%) on FrameNet 1.5 and FrameNet 1.7}
    \begin{threeparttable}
    \begin{adjustbox}{width=\linewidth}
    \begin{tabular}{lccc}
        \toprule
        \textbf{Approach} & \textbf{Accuracy} & \textbf{Precision} & \textbf{Recall} \\
        \midrule
        \multicolumn{4}{c}{\textbf{FrameNet 1.5}} \\
        \midrule
        KGFI (2021) \citep{su2021knowledge}           &      92     & - & 86   \\
        Tamburini  \citep{tamburini2022combining}     & \textbf{93} & - & -    \\
        COFFTEA \citep{an2023coarse}                  & \textbf{93} & - & 88   \\
        \midrule
        Baseline (Fine-Tuning without retrieval of candidates) & 24 & 34 & 43 \\
        RCIF (Zero-Shot) [without/with information about the number of Gold Frames]  & 12 / 13 & 12 / 18 & 50 / 33 \\
        RCIF (Few-Shot) [without/with information about the number of Gold Frames]  & 16 / 17 & 24 / 26 & 42 / 33 \\
        RCIF (Fine-Tuning) [without/with information about the number of Gold Frames]  & 89 / 92 & 91 / \textbf{92} & \textbf{92} / \textbf{92} \\
        \midrule
        \multicolumn{4}{c}{\textbf{FrameNet 1.7}} \\
        \midrule
        KGFI (2021) \citep{su2021knowledge}           &      92     & - & 86   \\
        Tamburini  \citep{tamburini2022combining}     &      92     & - & -    \\
        COFFTEA \citep{an2023coarse}                  &      93     & - & 87   \\
        \midrule
        Baseline (Fine-Tuning without retrieval of candidates) & 25 & 34 & 44 \\
        RCIF (Zero-Shot) [without/with information about the number of Gold Frames]  & 12 / 13 & 12 / 18 & 50 / 33 \\
        RCIF (Few-Shot) [without/with information about the number of Gold Frames]  & 16 / 17 & 24 / 26 & 42 / 33 \\
        RCIF (Fine-Tuning) [without/with information about the number of Gold Frames]  & \textbf{95} / 94 & \textbf{99} / 96 & \textbf{97} / \textbf{97} \\
        \bottomrule
    \end{tabular}
    \end{adjustbox}
    \end{threeparttable}
    \label{tab:frame-detection-results-fn15-fn17}
\end{table*}

\begin{table*}[h]
    \centering
    \caption{Performance (\%) on FrameNet 1.5 and FrameNet 1.7}
    \begin{adjustbox}{width=\textwidth}
    \begin{tabular}{lcccccc}
        \toprule
        \textbf{Approach} & \multicolumn{3}{c}{\textbf{FrameNet 1.5}} & \multicolumn{3}{c}{\textbf{FrameNet 1.7}} \\
        \cmidrule(lr){2-4} \cmidrule(lr){5-7}
         & \textbf{Recall@1} & \textbf{Recall@3} & \textbf{Recall@5} & \textbf{Recall@1} & \textbf{Recall@3} & \textbf{Recall@5} \\
        \midrule
        KGFI (2021) & 86 & - & - & 86 & - & - \\
        COFFTEA & 88 & 93 & 95 & 87 & 93 & 94 \\
        RCIF (Fine-Tuning) [without Info on \# Gold Frames] & 89 & 94 & 95 & 90 & 95 & 96 \\
        \bottomrule
    \end{tabular}
    \end{adjustbox}
    \label{tab:performance_fn}
\end{table*}

\begin{table*}
    \centering
    \caption{Performance (\%) with and without training on exemplars data}
    \begin{threeparttable}
    \begin{adjustbox}{width=\textwidth}
    \begin{tabular}{lccc}
        \toprule
        \textbf{Metrics} & \textbf{Accuracy} & \textbf{Precision} & \textbf{Recall} \\
        \midrule
        \multicolumn{4}{c}{\textbf{Training with just the training set}} \\
        \midrule
        FrameNet 1.5      & 44 & 52 & 44 \\
        FrameNet 1.7      & 48 & 49 & 49 \\
        \midrule
        \multicolumn{4}{c}{\textbf{Initial training using "exemplars" data then continue training on training set - Test performed using the testset}} \\
        \midrule
        FrameNet 1.5      & 87 & 89 & 88 \\
        FrameNet 1.7      & 92 & 94 & 95 \\
        \bottomrule
    \end{tabular}
    \end{adjustbox}
    \end{threeparttable}
    \label{tab:frame-detection-results-exemplars}
\end{table*}

Consistent with findings from \citep{chanin2023open} and \citep{an2023coarse}, our experiments using exemplar data for training—while testing on the same test split as in previous experiments—showed lower performance, as indicated in Table \ref{tab:frame-detection-results-exemplars}. Additionally, when exemplars are used as initial training data (i.e., training with exemplars first and then continuing with the official training split of the two datasets), the performance remains slightly lower compared to not using exemplars at all. We hypothesize that this drop in performance stems from the nature of exemplar data, where each sentence typically evokes only one frame, with annotations limited to a single frame per sentence. This characteristic makes exemplar data less suitable for training models intended to detect multiple frames per sentence, as previously noted by \citep{chanin2023open}.

\subsection{SPARQL query generation using LCQ2F and LCQ2F+} 
In this section we report the LLMs' performance on \textit{LCQ2F} and \textit{LCQ2F+} datasets introduced in Section \ref{sec:datasets-lcq2f-and-lcq2f+-construction}.

\begin{figure*}
\centering
\begin{minipage}[t]{1\textwidth}
    \centering
    \includegraphics[width=\linewidth]{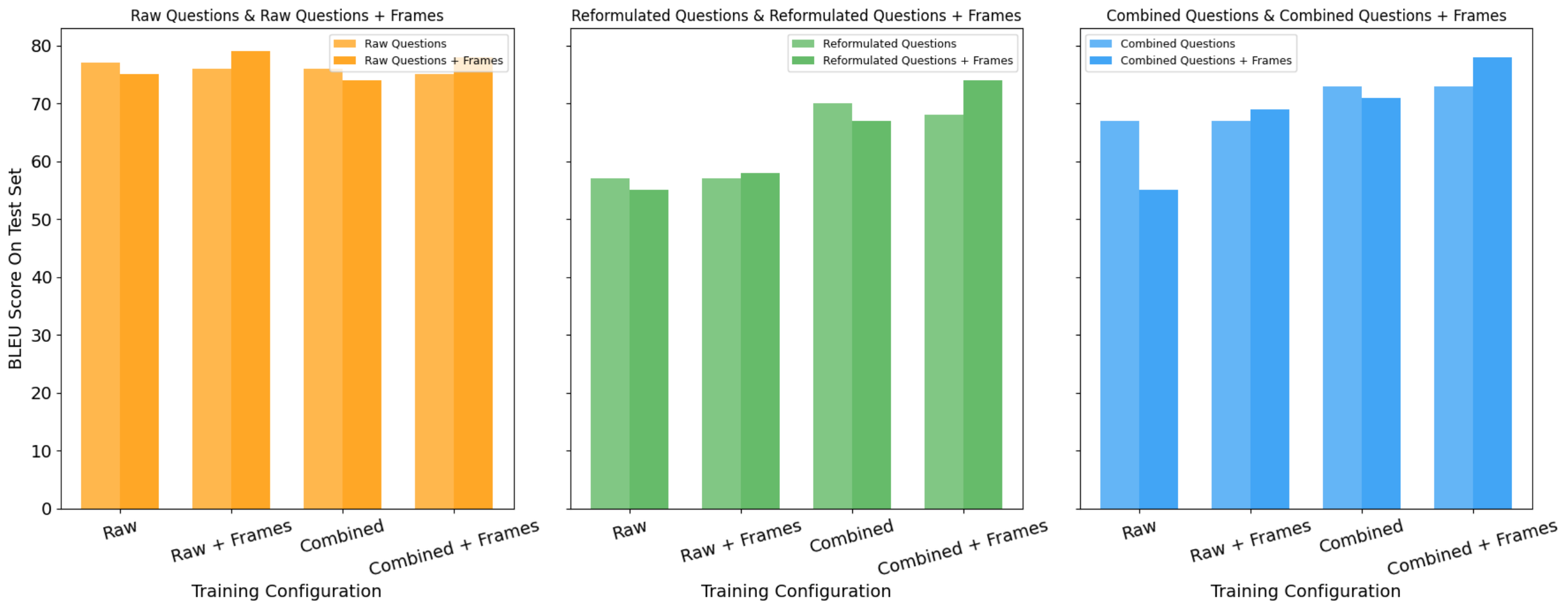}
    \caption{Llama3.1 BLEU Score Performances on LCQ2F}
    \label{fig:chart-llama3-results-lcq2f}
\end{minipage}
\end{figure*}

To obtain a comprehensive understanding of the model's behavior, we consider multiple training and testing configurations. At each stage, we evaluate two training setups: (1) training on template-based questions (\textit{raw questions}), and (2) training on a dataset that combine template-based and reformulated questions as separate entries, each mapped to the same SPARQL query (\textit{combined questions}). During testing, we introduce a third configuration (\textit{reformulated questions}), which assesses the model’s generalization ability towards natural language questions that were not generated using templates.
Each configuration was further evaluated with and without the inclusion of frame-based structured representations. This experimental design enabled us to measure the impact of incorporating structured representations alongside unstructured data. The results of these experiments are presented in Tables \ref{tab:llama3-results-sparql-generation-LCQ2F}, \ref{tab:llama3-results-sparql-generation-LCQ2F+} and Figure \ref{fig:chart-llama3-results-lcq2f}. Values in gray represent results from cross-setting experiments, where either the training or testing data lacks frame augmentation. In contrast, values in black correspond to experiments conducted within the same setting, meaning that both the training and testing data either include frames or omit them entirely.

From these results, we draw the following key insights:\\
- \textbf{Impact of the structured representations.} 
Our analysis of the results indicates that, despite the relatively small size of our constructed datasets, the use of frame-structured representations enhances performance for both template-based (\textit{raw questions}) and \textit{reformulated questions}. These \textit{Reformulated questions} allow us to test the generalization ability of the model and its capability of handling different natural language formulations. Let's examine the following two configurations: (1) training on \textit{raw questions} and testing on \textit{reformulated questions}, and (2) training on \textit{combined questions} and testing on the same \textit{reformulated questions} again. A comparison of these two settings reveals a significant improvement of 13 BLEU points in generalization performance (57 BLEU vs. 70 BLEU). This improvement is attributed to the larger dataset resulting from the combination of the two type of questions. This results is further improved to 74 BLEU when frames are used, demonstrating the effectiveness of structured representations, as highlighted in the orange block of Table \ref{tab:llama3-results-sparql-generation-LCQ2F}. Incorporating \textit{combined questions} with frames leads to the best performance on combined questions (78 BLEU), which is similar to the performance of the model on raw questions, which are traditionally easier, as shown in Table \ref{tab:llama3-results-sparql-generation-LCQ2F}. \\

\newpage
- \textbf{Performance difference between LCQ2F and LCQ2F+.} Overall, the performance in \textit{LCQ2F} dataset outperforms the one obtained with \textit{LCQ2F+} by a margin of 3\%. This indicates that the additional details in \textit{LCQ2F+}, such as the descriptions of frames and their associated Frame-Elements, may have introduced more noise leading to more complexity that confused the model. 
    
\begin{table*}
    \centering
    \caption{Llama 3.1-8B BLEU-Score performances on different configurations of LCQ2F.}
    \begin{threeparttable}
    \begin{adjustbox}{width=\textwidth}
    \begin{tabular}{l|cc|cc}
        \toprule
        \diagbox{\textbf{Data used for testing ↓}}{\textbf{Data used for training →}} & \multicolumn{2}{c|}{\textbf{Raw questions}} & \multicolumn{2}{c}{\textbf{Combined questions}} \\
        \midrule
        \midrule
        & - & with Frames & - & with Frames \\       
        \midrule
        
        Raw questions                      & \cellcolor{cyan!15}\textcolor{black}{\underline{77}} & \cellcolor{cyan!15}\textcolor{gray}{76} & \cellcolor{cyan!15}\textcolor{black}{76} & \cellcolor{cyan!15}\textcolor{gray}{75}  \\
        Raw questions with Frames          & \cellcolor{cyan!15}\textcolor{gray}{75} & \cellcolor{cyan!15}\textcolor{black}{\underline{79}} & \cellcolor{cyan!15}\textcolor{gray}{74} & \cellcolor{cyan!15}\textcolor{black}{78} \\
        \midrule
        
        Reformulated questions             & \cellcolor{orange!15}\textcolor{black}{57} & \cellcolor{orange!15}\textcolor{gray}{57} & \cellcolor{orange!15}\textcolor{black}{\underline{70}} & \cellcolor{orange!15}\textcolor{gray}{68}  \\
        Reformulated questions with Frames & \cellcolor{orange!15}\textcolor{gray}{55} & \cellcolor{orange!15}\textcolor{black}{58} & \cellcolor{orange!15}\textcolor{gray}{67} & \cellcolor{orange!15}\textcolor{black}{\underline{74}} \\
        \midrule
        
        Combined questions                 & \cellcolor{green!15}\textcolor{black}{67} & \cellcolor{green!15}\textcolor{gray}{67} & \cellcolor{green!15}\textcolor{black}{\underline{73}} & \cellcolor{green!15}\textcolor{gray}{73}  \\
        Combined questions with Frames     & \cellcolor{green!15}\textcolor{gray}{55} & \cellcolor{green!15}\textcolor{black}{69} & \cellcolor{green!15}\textcolor{gray}{71} & \cellcolor{green!15}\textcolor{black}{\underline{78}} \\
        \bottomrule
    \end{tabular}
    \end{adjustbox}
     \begin{tablenotes}
         \item[-] Each row represents a test split of the dataset (raw, reformulated, and combined questions).
         \item[-] The columns represent the splits used for training, while the rows are those used during testing. For each \\ split, we have the original versions of the questions (left) and those augmented with frames (right).
         \item[-] For example, the top-right cells depict the experiment using \textit{Combined questions} during training and \\ two versions of \textit{Raw questions} during testing with two versions of each dataset (with or without frames)
         \item[-] The underlined value in each row represents the best performance during testing for each test split.
    \end{tablenotes}
    \end{threeparttable}
    \label{tab:llama3-results-sparql-generation-LCQ2F}
\end{table*}

\begin{table*}
    \centering
    \caption{Llama 3.1-8B BLEU-Score performances on different configurations of LCQ2F+.}
    \begin{threeparttable}
    \begin{adjustbox}{width=\textwidth}
    \begin{tabular}{l|cc|cc}
        \toprule
        \diagbox{\textbf{Data used for testing ↓}}{\textbf{Data used for training →}} & \multicolumn{2}{c|}{\textbf{Raw questions}} & \multicolumn{2}{c}{\textbf{Combined questions}} \\
        \midrule
        \midrule
        & - & with Frames & - & with Frames \\
        \midrule
        
        Raw questions                      & \cellcolor{cyan!15}\textcolor{black}{\underline{77}} & \cellcolor{cyan!15}\textcolor{gray}{74} & \cellcolor{cyan!15}\textcolor{black}{76} & \cellcolor{cyan!15}\textcolor{gray}{75} \\
        Raw questions with Frames          & \cellcolor{cyan!15}\textcolor{gray}{75} & \cellcolor{cyan!15}\textcolor{black}{\underline{77}} & \cellcolor{cyan!15}\textcolor{gray}{74} & \cellcolor{cyan!15}\textcolor{black}{78} \\
        \midrule
        
        Reformulated questions             & \cellcolor{orange!15}\textcolor{black}{57} & \cellcolor{orange!15}\textcolor{gray}{48} & \cellcolor{orange!15}\textcolor{black}{\underline{70}} & \cellcolor{orange!15}\textcolor{gray}{67} \\
        Reformulated questions with Frames & \cellcolor{orange!15}\textcolor{gray}{55} & \cellcolor{orange!15}\textcolor{black}{59} & \cellcolor{orange!15}\textcolor{gray}{67} & \cellcolor{orange!15}\textcolor{black}{\underline{73}} \\
        \midrule
        
        Combined questions                 & \cellcolor{green!15}\textcolor{black}{67} & \cellcolor{green!15}\textcolor{gray}{64} & \cellcolor{green!15}\textcolor{black}{\underline{73}} & \cellcolor{green!15}\textcolor{gray}{72} \\
        Combined questions with Frames     & \cellcolor{green!15}\textcolor{gray}{55} & \cellcolor{green!15}\textcolor{black}{67} & \cellcolor{green!15}\textcolor{gray}{71} & \cellcolor{green!15}\textcolor{black}{\underline{75}} \\
        \bottomrule
    \end{tabular}
    \end{adjustbox}
    \end{threeparttable}
    \label{tab:llama3-results-sparql-generation-LCQ2F+}
\end{table*}






\section{Conclusion}
In this paper, we introduced a novel approach called \textbf{RCIF} (\textbf{R}etrieve \textbf{C}andidates and \textbf{I}dentify \textbf{F}rames) to frame detection leveraging RAG models. Unlike previous SOTA methods, which rely on predefined spans within the input text for frame detection, our method operates solely on the input text sequence without requiring additional information about the target span. Our proposed pipeline consists of two components: a candidates retriever and a LLM that selects the correct frames from the retrieved set of candidates. This approach demonstrated improved performance over SOTA methods on FrameNet 1.5 and achieved significantly higher performance on FrameNet 1.7, which provides a larger training set. Consequently, our method is well-suited to real-world applications where only raw text is available, and specific spans for frame detection are not predefined.
Furthermore, we showed that such structured representations with frames can enhance generalization towards template-free questions, and despite being trained on a relatively small dataset. This result highlights a promising research direction for improving the generalization capabilities of models to handle lexical variations. 

\section*{Limitations}
One limitation of this study is the relatively small size of the constructed datasets used to evaluate the impact of frame-structured representations on the generalization of SPARQL query generation for reformulated questions. Future research should explore methods for developing larger, high-quality datasets, extending beyond English to include multilingual resources. Additionally, this study is limited to the use of "frozen-RAG" for the retrieval component. Exploring trained versions of RAG could be a promising direction for future research.

\section*{Acknowledgments}
We are grateful to the NSERC Discovery Grant Program, which has funded this research. The authors would also like to express their gratitude to Compute Canada (Calcul Quebec) for providing computational resources.

\bibliography{biblio}

\appendix
\section{Appendix}
\label{sec:appendix}

\subsection{Frame representations for candidates retrieval component}
\label{sec:appendix-retrieval-component}
\begin{figure}[t!]
\centering
\includegraphics[width=1\linewidth]{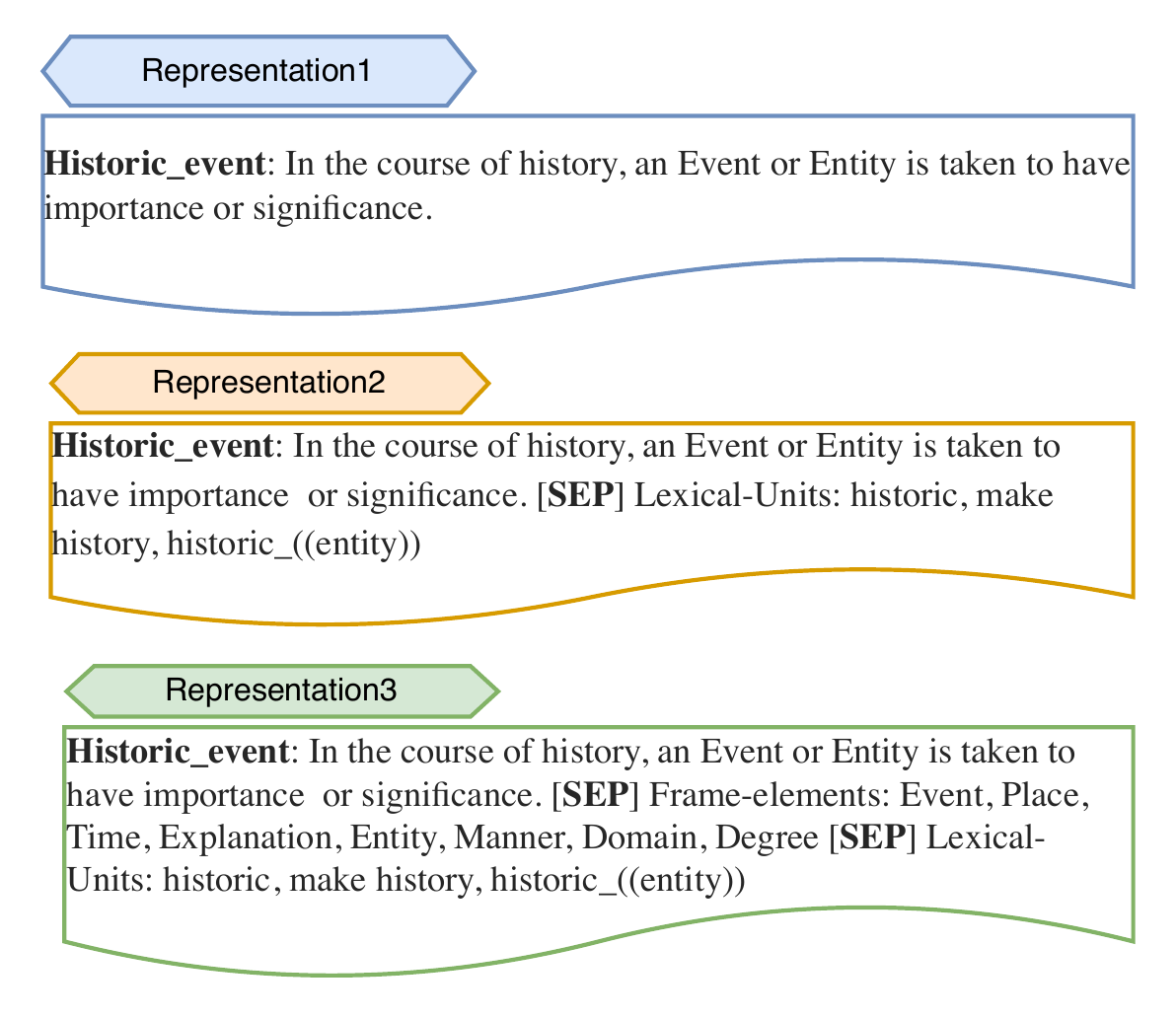}
\caption{Different representations of frames used in the retrieval component. \textit{Representation1} consists of the frame label and its textual description. \textit{Representation2} extends the previous one by appending a list of lexical units, while \textit{Representation3} further enriches \textit{Representation2} by incorporating a list of frame elements, resulting in a more comprehensive one.}
\label{fig:example_frames_representation}
\end{figure}

For the retrieval component, we compare different frame representations using labels, descriptions, lexical units (LUs), or frame elements (FEs).  Figure \ref{fig:example_frames_representation} shows different representations of the frame "Historic\_event". 

\subsection{Details about the frames detection component}
\label{sec:appendix-frame-detection-component}
\begin{figure}[t!]
\centering
\includegraphics[width=1\linewidth]{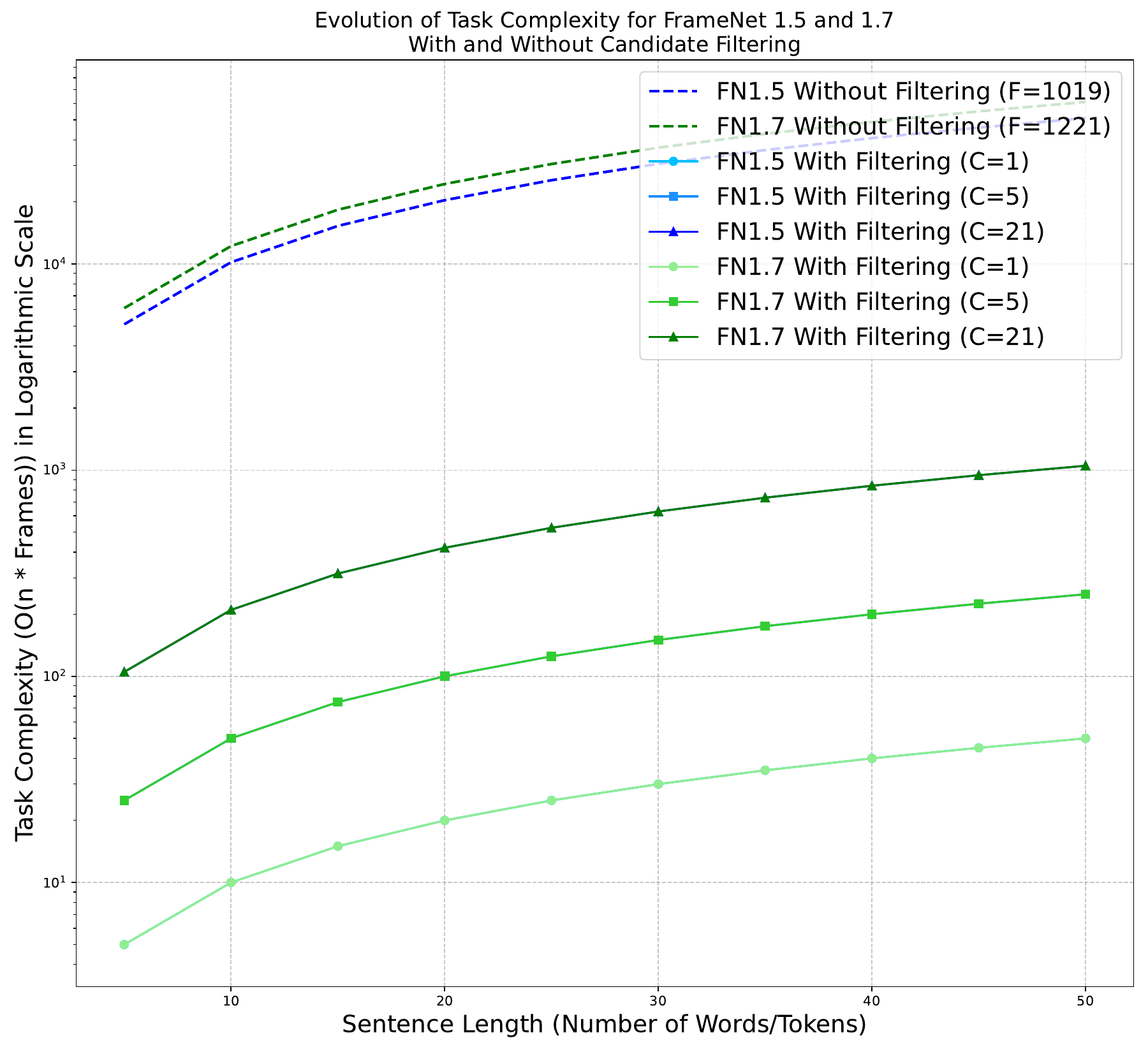}
\caption{Complexity evolution of the task of frame detection with and without candidates filtering and different values for the number of candidates \textit{C}.}
\label{fig:complexity_analysis_frame_detection}
\end{figure}

\begin{figure}[t!]
\centering
\includegraphics[width=1\linewidth]{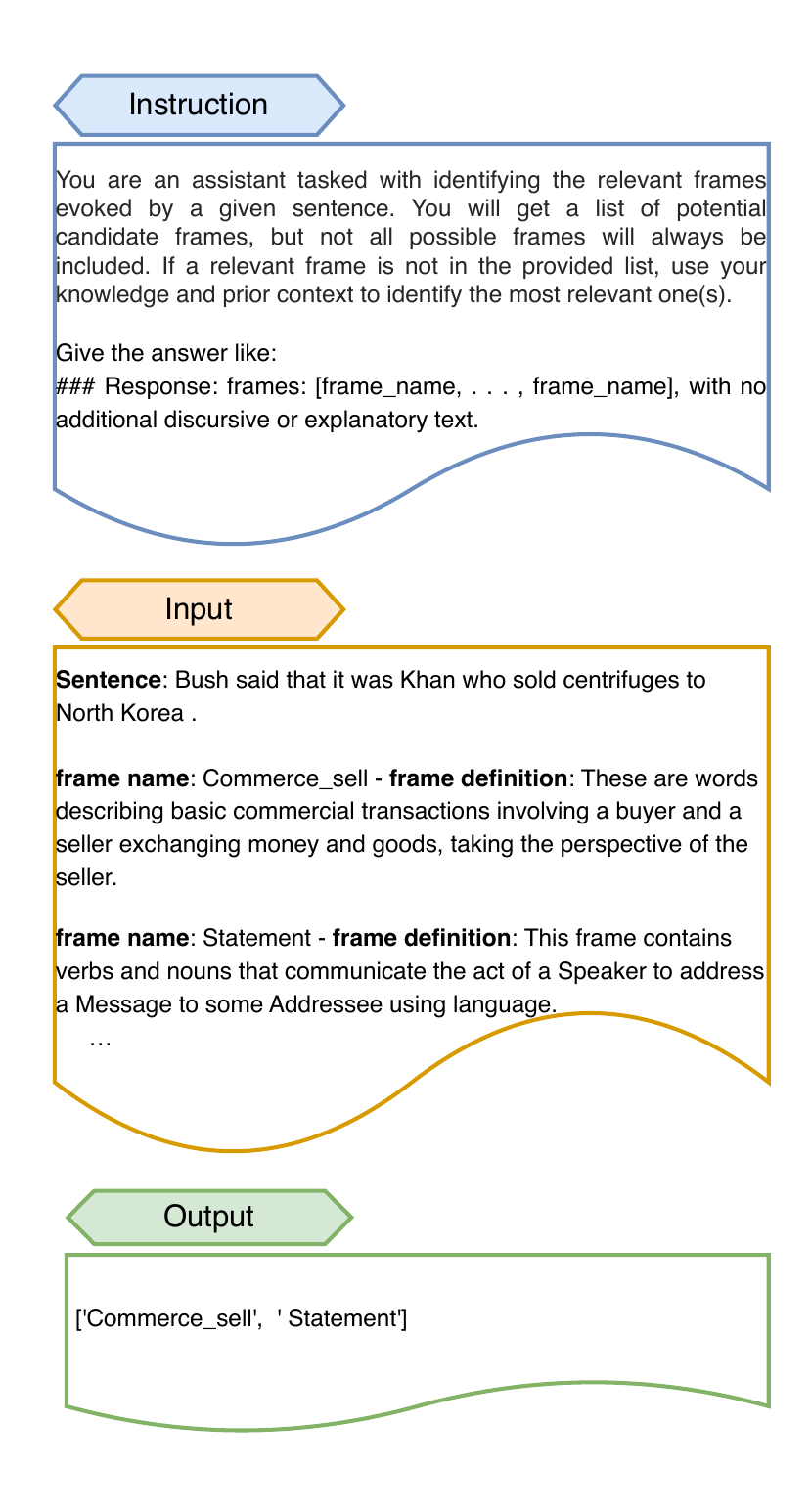}
\caption{Example of the dynamic prompt used to fine-tune the LLM}
\label{fig:example-prompt-for-fine-tuning}
\end{figure}

As depicted in Figure \ref{fig:complexity_analysis_frame_detection}, 
To reduce the complexity of frame detection (shown in \ref{fig:complexity_analysis_frame_detection}) using a Large Language Model (LLM), the input comprises a sentence alongside a list of potential candidate frames that could be evoked by the sentence, as illustrated in the input of Figure~\ref{fig:example-prompt-for-fine-tuning}. For each candidate frame, we provide its textual description, the list of Frame Elements (FEs), and the list of Lexical Units (LUs) associated with it. During training, the list of candidate frames varies dynamically from sample to sample. Consequently, a frame encountered in earlier iterations may not be included as a candidate in the current training instance, even if it is relevant to the sentence being processed. 

In such cases, it is crucial for the model to exhibit flexibility by "remembering" previously encountered frames and detecting them as evoked by the current sentence, provided they align with the context. To instill this capability, we designed the prompt as shown in the instruction of Figure~\ref{fig:example-prompt-for-fine-tuning}. This prompt design provides the following key benefits: (1) Leveraging the model's knowledge: By allowing the model to select frames beyond the given candidate list, this approach mitigates the risk of constraining the model to incomplete or imperfect candidate sets, thereby enhancing its flexibility and recall. (2) Generalization across frames: The model is trained to recognize associations between lexical units (LUs), Frame Elements (FEs), and frame descriptions across training instances. This enables it to generalize to detect relevant frames even when they are not explicitly listed among the current candidates. (3) Handling missing candidates: The prompt provides a mechanism to recover from situations where a relevant frame is inadvertently excluded from the candidate list. By instructing the model to infer and select the best-fitting frames when necessary, it ensures robust performance in scenarios with incomplete candidate information.

Using this technique, we successfully completed the list of candidate frames for 88\% of the samples with incomplete candidate sets. This result was obtained by evaluating, for each sample with an incomplete candidate list, the proportion of cases in which the model correctly identified the missing frame(s). The remaining 12\% are cases where at least one of guessed frames is irrelevant.

Technical details about the fine-tuning process performed using LoRA \citep{hu2021lora} 4-bit Quantization and SFTTrainer\footnote{https://huggingface.co/docs/trl/en/sft\_trainer} are provided in Table \ref{tab:fine-tuning-technical-details}. 

\begin{table}[h]
    \centering
    \caption{Technical details of the fine-tuning}
    \begin{threeparttable}
    \begin{adjustbox}{width=\linewidth}
    \begin{tabular}{lcc}
        \toprule
        \textbf{Parameters} & \textbf{Values}  \\
        \midrule
        GPU model & A100-40gb \\
        Number of hours (training and inference) & 12 \\
        Number of epochs & 10 \\
        Max Sequence Length & 2048 \\
        Packing & False (for faster training) \\
        Per Device Batch Size & 16 \\
        Gradient Accumulation Steps & 4 \\
        Warmup Steps & 5 \\
        Learning Rate & 2e-4 \\
        Precision Mode & fp16 or bf16 (conditional) \\
        Logging Steps & 1 \\
        Optimizer & adamw\_8bit \\
        Weight Decay & 0.01 \\
        Learning Rate Scheduler & Linear \\
        Random Seed & 3407 \\
        Evaluation Strategy & Epoch \\
        \bottomrule
    \end{tabular}
    \end{adjustbox}
    \end{threeparttable}
    \label{tab:fine-tuning-technical-details}
\end{table}

\subsection{Additional experiments with Phi-4 and Qwen2.5-7B-Instruct}
\label{sec:appendix-additional-experiments-phi4-qwen}
For our additional experiments with alternative large language models (LLMs), we employed two advanced models: Phi-4\footnote{\url{https://huggingface.co/microsoft/phi-4}} and Qwen2.5-7B-Instruct\footnote{\url{https://huggingface.co/Qwen/Qwen2.5-7B-Instruct}}. Phi-4 is a 14B parameter dense decoder-only Transformer model, designed with a focus on high-quality data and advanced reasoning. It was trained on a blend of synthetic datasets, filtered public domain content, and academic resources, ensuring a strong foundation for instruction adherence and safety. Through supervised fine-tuning and direct preference optimization, Phi-4 achieves precise alignment and robust instruction-following capabilities. In contrast, Qwen2.5-7B-Instruct is part of the Qwen2.5 model series, featuring enhanced knowledge, stronger coding and mathematical capabilities, and improved instruction following. It generates structured outputs such as JSON, and excels in handling structured data like tables which is interesting for SPARQL query generation.

\begin{figure*}[t!]
\centering
\begin{minipage}[t]{1\textwidth}
    \centering
    \includegraphics[width=\linewidth]{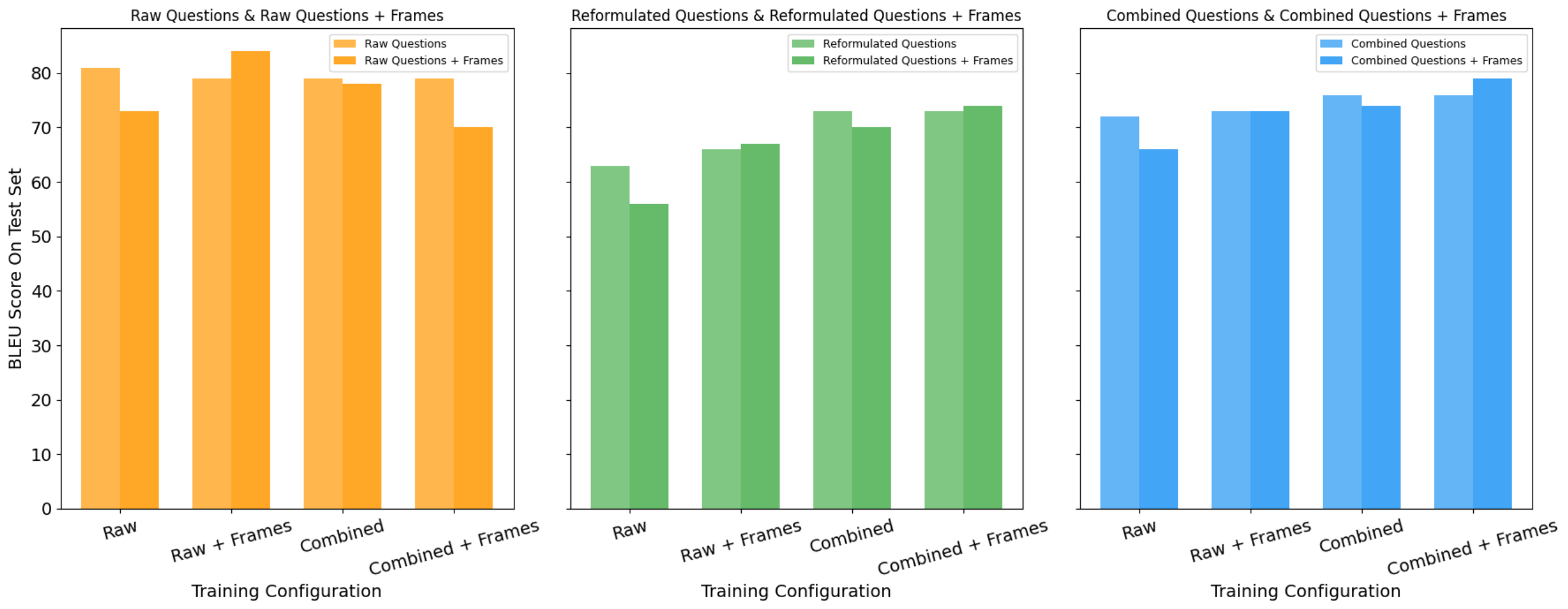}
    \caption{Phi-4 BLEU Score Performances on LCQ2F}
    \label{fig:chart-phi4-results-lcq2f}
\end{minipage}
\end{figure*}

\begin{figure*}[t!]
\centering
\begin{minipage}[t]{1\textwidth}
    \centering
    \includegraphics[width=\linewidth]{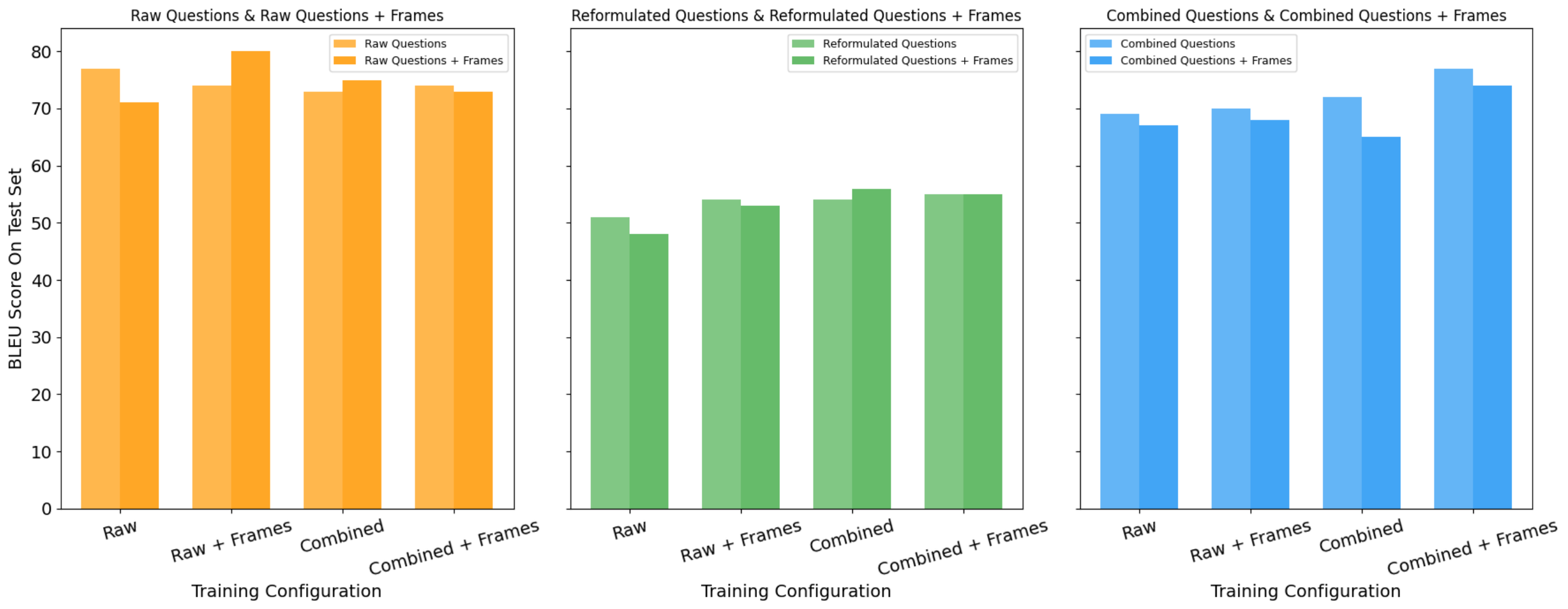}
    \caption{Qwen2.5-7B-Instruct BLEU Score Performances on LCQ2F}
    \label{fig:chart-qwen-results-lcq2f}
\end{minipage}
\end{figure*}

The results presented in Tables \ref{tab:phi4-results-sparql-generation-LCQ2F}, \ref{tab:phi4-results-sparql-generation-LCQ2F+}, \ref{tab:qwen_results-sparql-generation-LCQ2F} and \ref{tab:qwen_results-sparql-generation-LCQ2F+} as well as Figures \ref{fig:chart-phi4-results-lcq2f} and \ref{fig:chart-qwen-results-lcq2f} for the Phi-4 and Qwen models corroborate the findings observed in Tables \ref{tab:llama3-results-sparql-generation-LCQ2F} and \ref{tab:llama3-results-sparql-generation-LCQ2F+} with the Llama 3.2-3B model. Specifically, these results reaffirm the initial conclusions drawn from experiments with Llama 3.2-3B, namely: (1) Augmenting questions with frame-structured representations enhances generalization performance. (2) A structured representation based solely on frames yields better generalization than an augmentation strategy that incorporates textual descriptions of the evoked frames within the question.

Integrating both raw and reformulated questions into a larger dataset enhances generalization capabilities, particularly when questions are augmented with frame-structured representations. To illustrate this effect, we consider the dataset \textit{LCQ2F} (Tables \ref{tab:phi4-results-sparql-generation-LCQ2F} and \ref{tab:qwen_results-sparql-generation-LCQ2F}) and analyze the performance of models trained on either \textit{Raw Questions} or \textit{Combined Questions} when evaluated on the test set of the \textit{reformulated question} . Since these test entries are absent from both training splits, this comparison provides an unbiased assessment of generalization. On average, the Phi-4 model trained on \textit{Raw Questions} achieves a generalization performance of \( 63\% \) with reformulated questions, computed as \(\frac{63+66+56+67}{4} = 63\%\). In contrast, when trained on \textit{Combined Questions}, its generalization performance improves to \( 72.5\% \), calculated as \(\frac{73+73+70+74}{4} = 72.5\%\).

Furthermore, when analyzing generalization of Phi-4 over \textit{reformulated questions} (orange block in Table \ref{tab:phi4-results-sparql-generation-LCQ2F}) across different settings of training (model trained using questions without frames and and the one trained using questions with frames), we observe an average BLEU score of \(\frac{63+56}{2} = 59.5\) for questions without frames and \(\frac{66+67}{2} = 66.5\) for questions with frames. This indicates that incorporating frames improves the BLEU score by 7 points for \textit{Raw questions}. 

For \textit{Combined questions}, the inclusion of frames results in a 2-point BLEU score improvement, achieving the highest performance among \textit{reformulated questions} with a BLEU score of 74.

We observe the same phenomenon with Qwen2.5-7B-Instruct that has on average for \textit{Raw Questions} a BLEU score \(\frac{51+54+48+53}{4} = 51.5\%\) and a BLEU score for \textit{Combined Questions} being  \(\frac{54+55+56+55}{4} = 55\%\).

Once again, when evaluating the generalization of Qwen2.5-7B-Instruct on \textit{reformulated questions} (orange block in Table \ref{tab:qwen_results-sparql-generation-LCQ2F}) across different settings (questions without frames and questions with frames), we find that the average BLEU score is \(\frac{51+48}{2} = 49.5\) for questions without frames and \(\frac{54+53}{2} = 53.5\) for questions with frames. This demonstrates that incorporating frames leads to a 4-point improvement in BLEU score for \textit{Raw questions}.

\begin{table}[!ht]
    \centering
    \caption{Phi-4 BLEU-Score performances on different configurations of LCQ2F.}
    \begin{threeparttable}
    \begin{adjustbox}{width=\linewidth}
    \begin{tabular}{l|cc|cc}
        \toprule
        \diagbox{\textbf{Testing ↓}}{\textbf{Training →}} & \multicolumn{2}{c|}{\textbf{Raw Qst}} & \multicolumn{2}{c}{\textbf{Combined Qst}} \\
        \midrule
        \midrule
        & - & with Frames & - & with Frames \\
        \midrule
        
        Raw Qst                      & \cellcolor{cyan!15}\textcolor{black}{\underline{81}} & \cellcolor{cyan!15}\textcolor{gray}{79} & \cellcolor{cyan!15}\textcolor{black}{79} & \cellcolor{cyan!15}\textcolor{gray}{79} \\
        
        Raw Qst + Frames & \cellcolor{cyan!15}\textcolor{gray}{73} & \cellcolor{cyan!15}\textcolor{black}{\underline{84}} & \cellcolor{cyan!15}\textcolor{gray}{78} & \cellcolor{cyan!15}\textcolor{black}{70} \\
        \midrule
        
        Reformulated Qst             & \cellcolor{orange!15}\textcolor{black}{63} & \cellcolor{orange!15}\textcolor{gray}{66} & \cellcolor{orange!15}\textcolor{black}{\underline{73}} & \cellcolor{orange!15}\textcolor{gray}{\underline{73}} \\
        
        Reformulated Qst + Frames & \cellcolor{orange!15}\textcolor{gray}{56} & \cellcolor{orange!15}\textcolor{black}{67} & \cellcolor{orange!15}\textcolor{gray}{70} & \cellcolor{orange!15}\textcolor{black}{\underline{74}} \\
        \midrule
        
        Combined Qst                 & \cellcolor{green!15}\textcolor{black}{72} & \cellcolor{green!15}\textcolor{gray}{73} & \cellcolor{green!15}\textcolor{black}{\underline{76}} & \cellcolor{green!15}\textcolor{gray}{\underline{76}} \\
        
        Combined Qst + Frames     & \cellcolor{green!15}\textcolor{gray}{66} & \cellcolor{green!15}\textcolor{black}{73} & \cellcolor{green!15}\textcolor{gray}{74} & \cellcolor{green!15}\textcolor{black}{\underline{79}} \\
        \bottomrule
    \end{tabular}
    \end{adjustbox}
    \begin{tablenotes}
         \item[•] The underlined value in each row indicates the hi-\\ghest performance achieved during testing for the \\ corresponding configuration.
    \end{tablenotes}
    \end{threeparttable}
    \label{tab:phi4-results-sparql-generation-LCQ2F}
\end{table}

\begin{table}[!ht]
    \centering
    \caption{Phi-4 BLEU-Score performances on different configurations of LCQ2F+.}
    \begin{threeparttable}
    \begin{adjustbox}{width=\linewidth}
    \begin{tabular}{l|cc|cc}
        \toprule
        \diagbox{\textbf{Testing ↓}}{\textbf{Training →}} & \multicolumn{2}{c|}{\textbf{Raw Qst}} & \multicolumn{2}{c}{\textbf{Combined Qst}} \\
        \midrule
        \midrule
        & - & with Frames & - & with Frames \\
        \midrule
        
        Raw Qst                      & \cellcolor{cyan!15}\textcolor{black}{\underline{81}} & \cellcolor{cyan!15}\textcolor{gray}{79} & \cellcolor{cyan!15}\textcolor{black}{77} & \cellcolor{cyan!15}\textcolor{gray}{75} \\
        
        Raw Qst + Frames          & \cellcolor{cyan!15}\textcolor{gray}{78} & \cellcolor{cyan!15}\textcolor{black}{70} & \cellcolor{cyan!15}\textcolor{gray}{76} & \cellcolor{cyan!15}\textcolor{black}{\underline{80}} \\
        \midrule
        
        Reformulated Qst             & \cellcolor{orange!15}\textcolor{black}{\underline{56}} & \cellcolor{orange!15}\textcolor{gray}{55} & \cellcolor{orange!15}\textcolor{black}{54} & \cellcolor{orange!15}\textcolor{gray}{55} \\
        
        Reformulated Qst + Frames & \cellcolor{orange!15}\textcolor{gray}{52} & \cellcolor{orange!15}\textcolor{black}{58} & \cellcolor{orange!15}\textcolor{gray}{52} & \cellcolor{orange!15}\textcolor{black}{\underline{60}} \\
        \midrule
        
        Combined Qst                 & \cellcolor{green!15}\textcolor{black}{65} & \cellcolor{green!15}\textcolor{gray}{68} & \cellcolor{green!15}\textcolor{black}{\underline{76}} & \cellcolor{green!15}\textcolor{gray}{\underline{75}} \\
        
        Combined Qst + Frames     & \cellcolor{green!15}\textcolor{gray}{63} & \cellcolor{green!15}\textcolor{black}{67} & \cellcolor{green!15}\textcolor{gray}{65} & \cellcolor{green!15}\textcolor{black}{\underline{78}} \\
        \bottomrule
    \end{tabular}
    \end{adjustbox}
    \end{threeparttable}
    \label{tab:phi4-results-sparql-generation-LCQ2F+}
\end{table}

\begin{table}[!ht]
    \centering
    \caption{Qwen2.5-7B-Instruct BLEU-Score performances on different configurations of LCQ2F.}
    \begin{threeparttable}
    \begin{adjustbox}{width=\linewidth}
    \begin{tabular}{l|cc|cc}
        \toprule
        \diagbox{\textbf{Testing ↓}}{\textbf{Training →}} & \multicolumn{2}{c|}{\textbf{Raw Qst}} & \multicolumn{2}{c}{\textbf{Combined Qst}} \\
        \midrule
        \midrule
        & - & with Frames & - & with Frames \\
        \midrule
        
        Raw Qst                      & \cellcolor{cyan!15}\textcolor{black}{\underline{77}} & \cellcolor{cyan!15}\textcolor{gray}{74} & \cellcolor{cyan!15}\textcolor{black}{73} & \cellcolor{cyan!15}\textcolor{gray}{74} \\
        
        Raw Qst + Frames          & \cellcolor{cyan!15}\textcolor{gray}{71} & \cellcolor{cyan!15}\textcolor{black}{\underline{80}} & \cellcolor{cyan!15}\textcolor{gray}{75} & \cellcolor{cyan!15}\textcolor{black}{73} \\
        \midrule
        
        Reformulated Qst             & \cellcolor{orange!15}\textcolor{black}{51} & \cellcolor{orange!15}\textcolor{gray}{54} & \cellcolor{orange!15}\textcolor{black}{54} & \cellcolor{orange!15}\textcolor{gray}{\underline{55}} \\
        
        Reformulated Qst + Frames & \cellcolor{orange!15}\textcolor{gray}{48} & \cellcolor{orange!15}\textcolor{black}{53} & \cellcolor{orange!15}\textcolor{gray}{\underline{56}} & \cellcolor{orange!15}\textcolor{black}{55} \\
        \midrule
        
        Combined Qst                 & \cellcolor{green!15}\textcolor{black}{69} & \cellcolor{green!15}\textcolor{gray}{70} & \cellcolor{green!15}\textcolor{black}{72} & \cellcolor{green!15}\textcolor{gray}{\underline{77}} \\
        
        Combined Qst + Frames     & \cellcolor{green!15}\textcolor{gray}{67} & \cellcolor{green!15}\textcolor{black}{68} & \cellcolor{green!15}\textcolor{gray}{65} & \cellcolor{green!15}\textcolor{black}{\underline{74}} \\
        \bottomrule
    \end{tabular}
    \end{adjustbox}
    \end{threeparttable}
    \label{tab:qwen_results-sparql-generation-LCQ2F}
\end{table}

\begin{table}[!ht]
    \centering
    \caption{Qwen2.5-7B-Instruct BLEU-Score performances on different configurations of LCQ2F+.}
    \begin{threeparttable}
    \begin{adjustbox}{width=\linewidth}
    \begin{tabular}{l|cc|cc}
        \toprule
        \diagbox{\textbf{Testing ↓}}{\textbf{Training →}} & \multicolumn{2}{c|}{\textbf{Raw Qst}} & \multicolumn{2}{c}{\textbf{Combined Qst}} \\
        \midrule
        \midrule
        & - & with Frames & - & with Frames \\
        \midrule
        
        Raw Qst                      & \cellcolor{cyan!15}\textcolor{black}{\underline{77}} & \cellcolor{cyan!15}\textcolor{gray}{75} & \cellcolor{cyan!15}\textcolor{black}{74} & \cellcolor{cyan!15}\textcolor{gray}{\underline{77}} \\
        
        Raw Qst + Frames          & \cellcolor{cyan!15}\textcolor{gray}{\underline{65}} & \cellcolor{cyan!15}\textcolor{black}{60} & \cellcolor{cyan!15}\textcolor{gray}{67} & \cellcolor{cyan!15}\textcolor{black}{67} \\
        \midrule
        
        Reformulated Qst             & \cellcolor{orange!15}\textcolor{black}{53} & \cellcolor{orange!15}\textcolor{gray}{54} & \cellcolor{orange!15}\textcolor{black}{\underline{56}} & \cellcolor{orange!15}\textcolor{gray}{53} \\
        
        Reformulated Qst + Frames & \cellcolor{orange!15}\textcolor{gray}{55} & \cellcolor{orange!15}\textcolor{black}{56} & \cellcolor{orange!15}\textcolor{gray}{57} & \cellcolor{orange!15}\textcolor{black}{\underline{58}} \\
        \midrule
        
        Combined Qst                 & \cellcolor{green!15}\textcolor{black}{57} & \cellcolor{green!15}\textcolor{gray}{59} & \cellcolor{green!15}\textcolor{black}{\underline{72}} & \cellcolor{green!15}\textcolor{gray}{\underline{58}} \\
        
        Combined Qst + Frames     & \cellcolor{green!15}\textcolor{gray}{61} & \cellcolor{green!15}\textcolor{black}{68} & \cellcolor{green!15}\textcolor{gray}{\underline{67}} & \cellcolor{green!15}\textcolor{black}{65} \\
        \bottomrule
    \end{tabular}
    \end{adjustbox}
    \end{threeparttable}
    \label{tab:qwen_results-sparql-generation-LCQ2F+}
\end{table}

\subsection{Frame Semantic Role Labeling (FSRL)}
\label{sec:appendix-fsrl}
In this section, we present zero-shot experiments aimed at aligning arguments (text spans associated with elements of a frame) to frame elements within frames identified by our frame detection framework, \textbf{RCIF}. Figure \ref{fig:argument-fe_mapping} illustrates the input configuration provided to the LLM (Mistral 7B in this case) and the corresponding output generated in these experiments. Given a sentence and an evoked frame, we employ an LLM to map each frame element to its respective text span within the sentence. To assess the capabilities of the LLM in this task, we use the Mistral 7B Instruct v0.3 model \citep{jiang2023mistral7b} and the WikiBank dataset \citep{sas2020wikibank} as our data source. The Mistral model’s output, illustrated in Figure \ref{fig:argument-fe_mapping}, shows promising potential for semantic role labeling tasks. However, a notable challenge arises: when the original sentence is replaced by a semantically similar one, the model’s matching performance declines significantly, as depicted in Figure \ref{fig:argument-fe_mapping2}. This highlights the need for further refinement if such models are to be effectively applied to frame semantic role labeling tasks. Future research should therefore focus on addressing this limitation.

\paragraph{Illustration of prompt used to perform FSRL.}
Figures \ref{fig:argument-fe_mapping} and \ref{fig:argument-fe_mapping2} illustrate prompts used for role labeling by aligning frame elements with their corresponding text spans from the input. The output in Figure \ref{fig:argument-fe_mapping} demonstrates a correct alignment, whereas Figure \ref{fig:argument-fe_mapping2} highlights a case with incorrect matching.

\subsection{Overview of \textit{LC-QuAD 2.0} and \textit{WikiBank}}
\label{sec:introduction-of-lcq2-and-wikibank}
In this section we introduce the two foundational datasets, LC-QuAD2.0 \citep{LCQUAD2} and WikiBank \citep{sas2020wikibank}, as they form the basis for constructing the derived datasets, \textit{LCQ2F} and \textit{LCQ2F+}. \\
- \textbf{LC-QuAD2.0}: LC-QuAD 2.0 \citep{LCQUAD2} dataset is a widely used, publicly available benchmark, consisting of question-query pairs referred to as entries. Each entry comprises a question in English and its corresponding SPARQL query as the target. The dataset was generated automatically using a templates-based approach, where templates align question-query structures with placeholders, later replaced by specific URIs. These aligned pairs are referred to as global templates. 

    For instance, the global template:  
    \begin{quote}
        \textit{Question}: "What is the \texttt{<1>} of \texttt{<2>}?"  \\
        \textit{Query}: \texttt{SELECT DISTINCT ?uri WHERE \{<2> <1> ?uri\}}  
    \end{quote}
    
    produces the following entry:  
    
    \begin{quote}
        \textit{Entry Question}: "What is the country with the max individual tax rate?"  \\
        \textit{Entry Query}:  
        \texttt{SELECT ?ent WHERE \{ ?ent wdt:P31 wd:Q6256 . ?ent wdt:P2834 ?obj \} ORDER BY DESC (?obj) LIMIT 5}
    \end{quote}

    The LC-QuAD2.0 dataset comprises a total of 30,225 entries, with 21,761 allocated to the training set, 2,418 to the validation set, and 6,046 to the test set. \\
- \textbf{WikiBank}: WikiBank is a multilingual dataset constructed to enhance frame-semantic parsing by aligning semantic structures from Wikidata with sentences from Wikipedia. This resource includes partially annotated sentences in multiple languages, using a heuristic-based approach to extract semantic roles and argument structures from knowledge base triples. By leveraging distant supervision, WikiBank facilitates cross-lingual transfer and significantly improves semantic parsing performance, particularly in low-resource languages. The dataset has demonstrated its utility in training frame-semantic parsers by providing substantial improvements when integrated into existing parsing frameworks, such as SLING, in both monolingual and multilingual settings. WikiBank contains over 795,000 sentences and 990,000 examples, but it is limited to a set of 247 relations derived from Wikidata, characterized by a highly uneven distribution.
Although the LC-QuAD2.0 dataset provides the advantage of containing a comprehensive set of questions paired with their corresponding SPARQL queries, it lacks the frame-semantic parsing annotations available in WikiBank. To leverage the strengths of both resources, we construct new datasets by intersecting the two.
\subsection{Details about the constructed datasets \textit{LCQ2F} and \textit{LCQ2F+}}
\label{sec:appendix-constructed-datasets}
\begin{figure*}[t!]
\centering
\begin{minipage}[t]{0.48\textwidth}
    \centering
    \includegraphics[width=\linewidth]{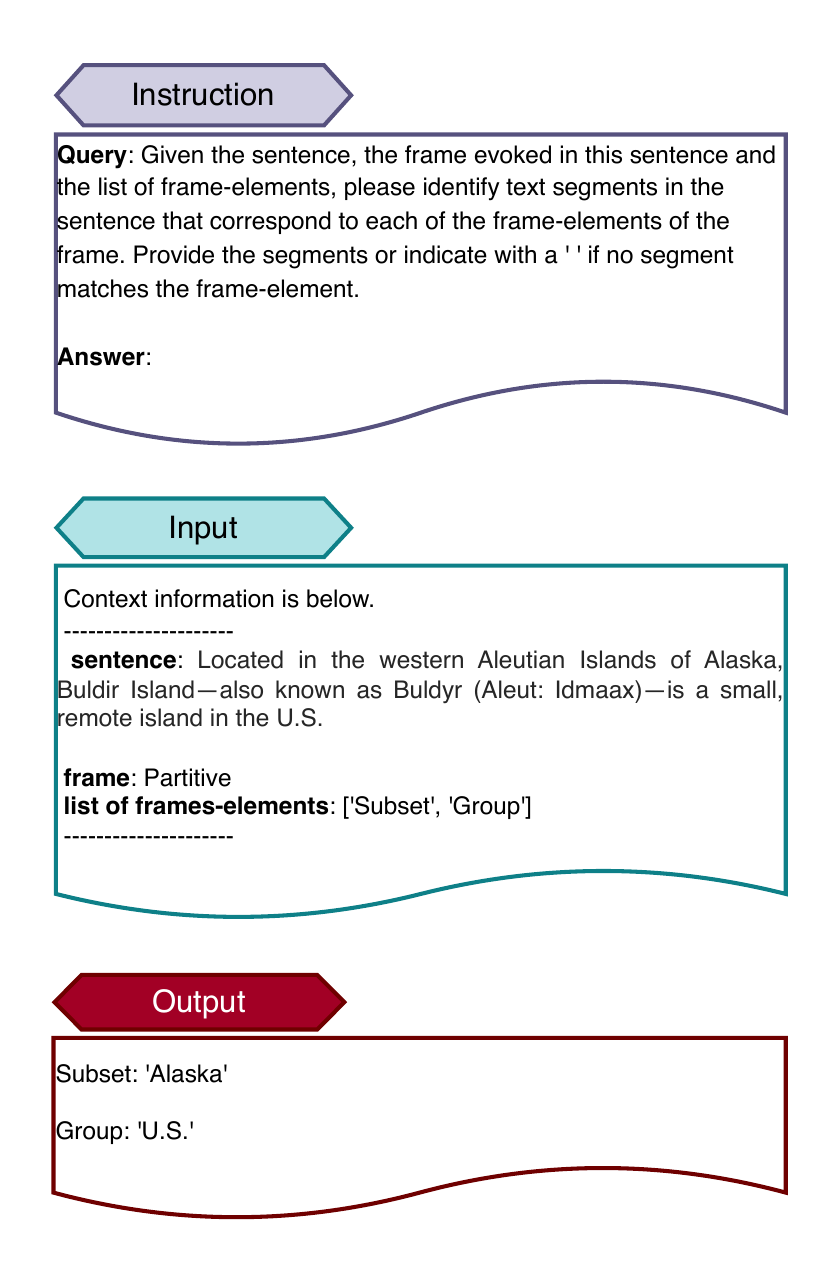}
    \caption{Example of the dynamic prompt used to perform role labeling with a good matching.}
    \label{fig:argument-fe_mapping}
\end{minipage}
\hfill
\begin{minipage}[t]{0.48\textwidth}
    \centering
    \includegraphics[width=\linewidth]{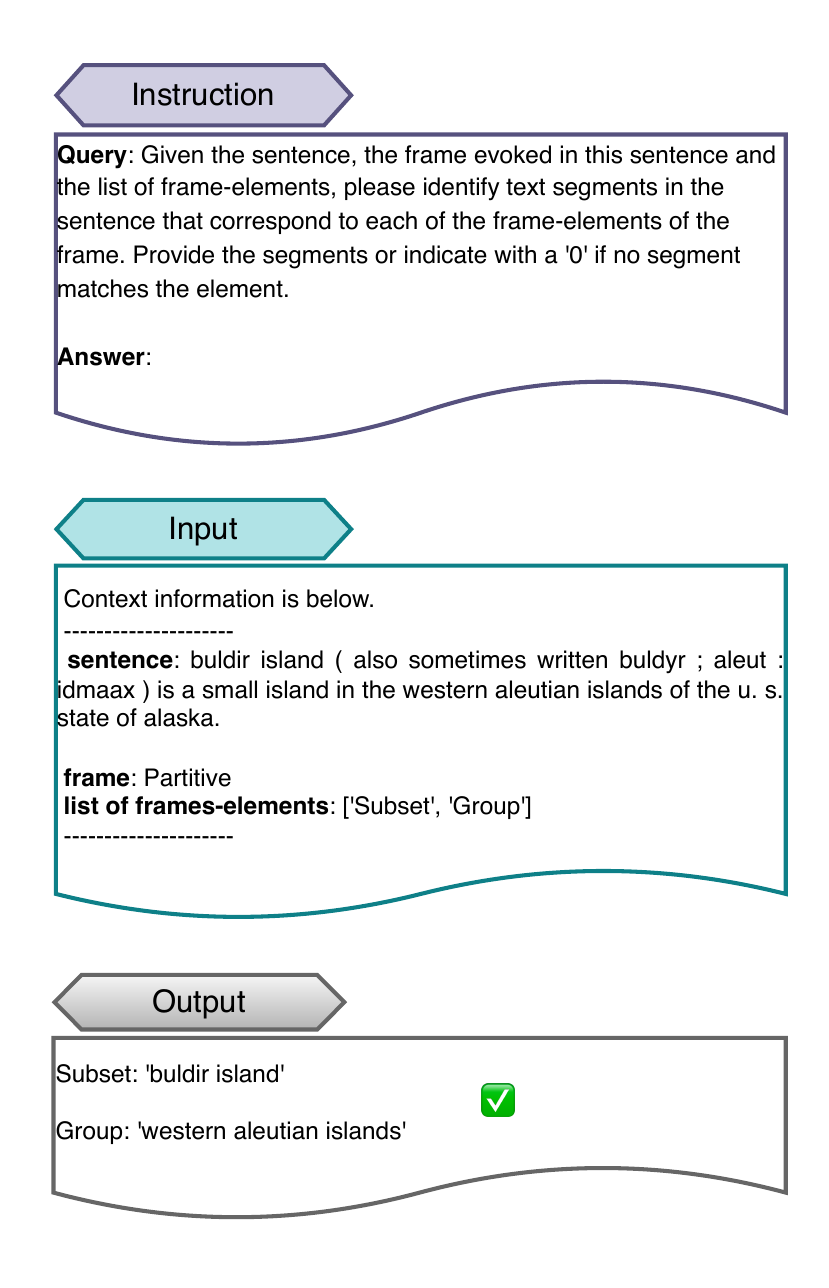}
    \caption{Example of the dynamic prompt used to perform role labeling with a wrong matching.}
    \label{fig:argument-fe_mapping2}
\end{minipage}

\vspace{0.5cm}

\includegraphics[width=\textwidth]{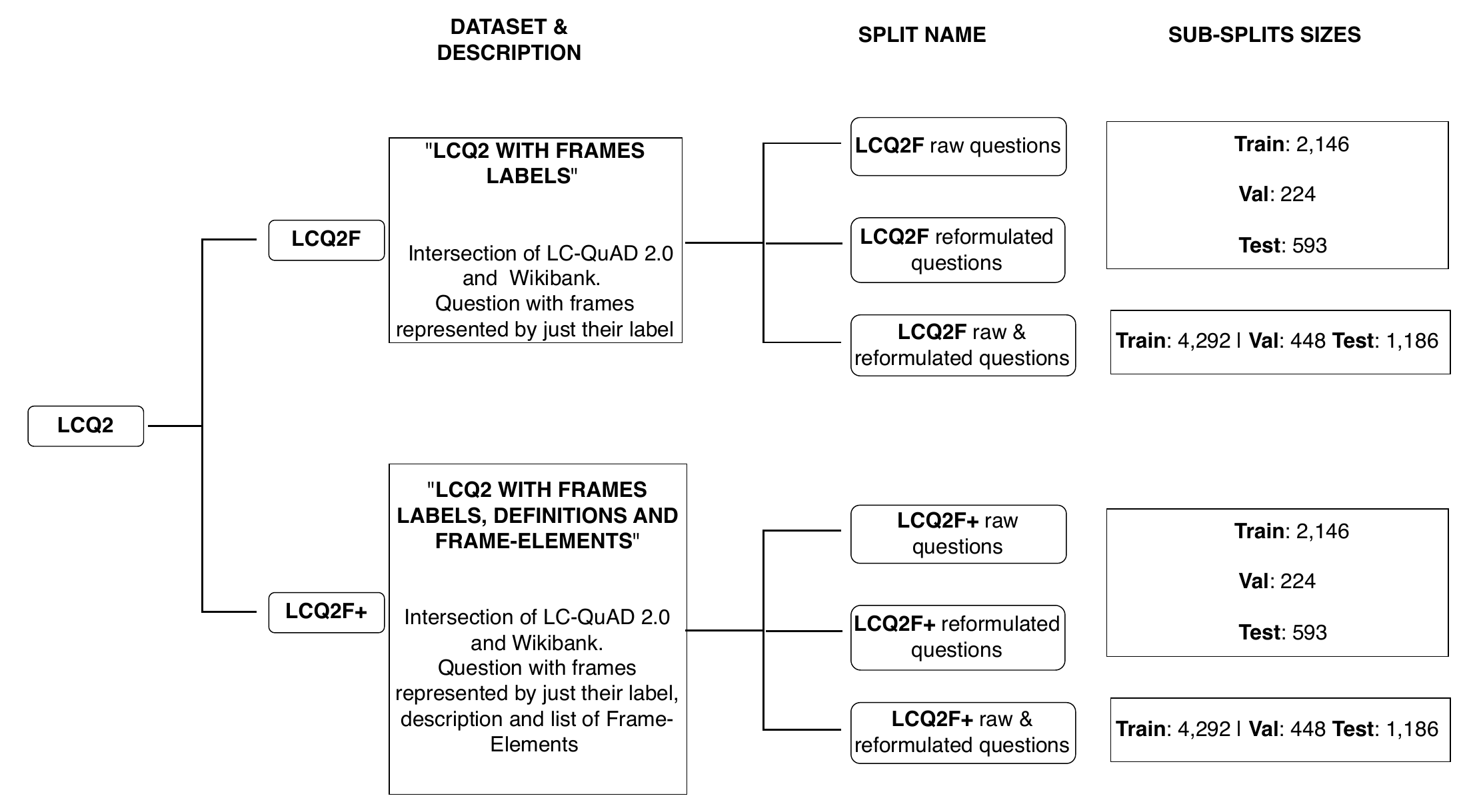}
\caption{Datasets descriptions.}
\label{fig:lcq2f_lcq2f+_datasets_description}
\end{figure*}

Figure \ref{fig:lcq2f_lcq2f+_datasets_description} provides a detailed description of the dataset derived from LC-QuAD 2.0 and WikiBank.

\end{document}